\documentclass[sigconf, accepted]{acmart}
\AtBeginDocument{%
  }

\setcopyright{acmlicensed}
\copyrightyear{2026}
\acmYear{2026}
\setcopyright{cc}
\setcctype{by}
\acmConference[DocEng '26]{Proceedings of the 2026 ACM Symposium on Document Engineering}{August 25--28, 2026}{Fribourg, Switzerland}
\acmBooktitle{Proceedings of the 2026 ACM Symposium on Document Engineering (DocEng '26), August 25--28, 2026, Fribourg, Switzerland}
\acmDOI{10.1145/3820755.3833407}
\acmISBN{979-8-4007-2786-3/2026/08}

\usepackage{hyperref}
\usepackage{multirow}
\usepackage{multicol}
\usepackage{pgfplots}
\usepackage[table]{xcolor}
\definecolor{highlight}{HTML}{A7D7D7}
\definecolor{plotcolour1}{HTML}{8dd3c7}
\definecolor{plotcolour2}{HTML}{fdb462}
\definecolor{plotcolour3}{HTML}{bebada}
\definecolor{plotcolour4}{HTML}{fb8072}
\definecolor{plotcolour5}{HTML}{80b1d3}
\definecolor{plotcolour6}{HTML}{ffd92f}
\definecolor{plotcolour7}{HTML}{6baed6}
\definecolor{faded}{HTML}{9e9e9e}
\colorlet{grey}{gray!60}

\pgfplotsset{compat=1.18}

\usepackage{pifont}
\definecolor{darkgreen}{HTML}{18864b}
\definecolor{darkred}{HTML}{bd451a}
\newcommand{\cmark}{\textcolor{darkgreen}{\ding{51}}}%
\newcommand{\xmark}{\textcolor{darkred}{\ding{55}}}%

\begin{document}

\title{The Price of Reasoning: Cost-Quality Tradeoffs in Reinforcement Learning for Neural Machine Translation}

\author{Michael Jungo}
\email{michael.jungo@unifr.ch}
\orcid{0009-0001-1790-1687}
\correspondingauthor
\affiliation{%
  \institution{AIBEX, University of Fribourg}
  \city{Fribourg}
  \country{Switzerland}
}

\author{Aixiu An}
\email{aixiu.an@hefr.ch}
\orcid{0000-0003-2919-1333}
\correspondingauthor
\affiliation{%
  \institution{iCoSys, University of Applied Sciences and Arts Western Switzerland}
  \city{Fribourg}
  \country{Switzerland}
}

\renewcommand{\shortauthors}{Jungo et al.}

\begin{abstract}
Reinforcement learning with verifiable rewards (RLVR) has been established as a
viable paradigm for the post-training of Large Language Models (LLMs), including
downstream tasks, such as Neural Machine Translation (NMT). With the latest
research indicating that RLVR could be the preferred training method for
translating legal documents due to the induced reasoning capabilities, it raises
the question whether it is really attributed to the reasoning or more generally
to the training paradigm. We investigate the importance of including the model's
reasoning trace in the generated responses during both \textit{training} and
\textit{inference} by systematically omitting it from one of the phases. Our
experiments show that including the reasoning, specifically during inference,
has a positive effect on the overall translation quality. Furthermore, we
recognise that the reasoning leads to an increase in output tokens, hence we
study the cost-quality tradeoff between the increased computational
demands and the improved translation quality.
\end{abstract}

\ccsdesc[500]{Computing methodologies~Machine translation}
\ccsdesc[500]{Computing methodologies~Reinforcement learning}
\ccsdesc[500]{Applied computing~Law}

\received{8 July 2026}
\received[accepted]{20 July 2026}

\maketitle

\begin{figure*}[ht]
    \centering
    \pgfplotslegendfromname{sharedlegend}
    \begin{tikzpicture}
        \begin{axis}[
            name=ax1,
            title={\textbf{Trained with Thinking [T:\cmark]}},
            title style={yshift=-4pt},
            width=0.48*\linewidth,
            height=6cm,
            xlabel={Number of Training Samples},
            ylabel={Mean COMET Score},
            legend style={
                at={(0.5, 1.05)},
                anchor=south, 
                legend columns=4,
                font=\small,
                draw=none,
                /tikz/every even column/.append style={column sep=0.5cm}
            },
            legend to name=sharedlegend,
            grid=major,
            grid style={gray!15, dashed},
            tick align=outside,
            xtick pos=bottom,
            ytick pos=left,
            x tick label style={
                /pgf/number format/.cd,
                set thousands separator={\,}
            },
            xtick distance=500,
            enlarge x limits=0.03,
            restrict x to domain=0:2500,
            ymin=74.5,
            ymax=83,
        ]
            \addplot+[
                mark=*,
                color=plotcolour1,
                mark options={fill=plotcolour1, draw=plotcolour1!90!black, solid},
                thick,
            ] table [x=num_samples, y expr=\thisrow{mean_comet}*100, col sep=comma] {every100-qwen3.5-4b.csv};
            
            \addplot+[
                mark=o,
                color=plotcolour1,
                mark options={fill=plotcolour1, draw=plotcolour1!90!black, solid},
                dashed,
                thick,
            ] table [x=num_samples, y expr=\thisrow{mean_comet}*100, col sep=comma] {every100-qwen3.5-4b-enable_thinking=False.csv};
            
            \addplot+[
                mark=*,
                color=plotcolour2,
                mark options={fill=plotcolour2, draw=plotcolour2!90!black, solid},
                thick,
            ] table [x=num_samples, y expr=\thisrow{mean_comet}*100, col sep=comma] {every100-qwen3.5-9b.csv};
            
            \addplot+[
                mark=o,
                color=plotcolour2,
                mark options={fill=plotcolour2, draw=plotcolour2!90!black, solid},
                dashed,
                thick,
            ] table [x=num_samples, y expr=\thisrow{mean_comet}*100, col sep=comma] {every100-qwen3.5-9b-enable_thinking=False.csv};
        \end{axis}
        
        \begin{axis}[
            name=ax2,
            at={(ax1.right of origin)},
            anchor=left of origin,
            title={\textbf{Trained without Thinking [T:\xmark]}},
            title style={yshift=-4pt},
            width=0.48*\linewidth,
            height=6cm,
            xshift=1.0cm,
            xlabel={Number of Training Samples},
            legend style={
                at={(0.5, 1.05)},
                anchor=south, 
                legend columns=4,
                font=\small,
                draw=none,
                /tikz/every even column/.append style={column sep=0.5cm}
            },
            legend to name=sharedlegend,
            grid=major,
            grid style={gray!15, dashed},
            tick align=outside,
            xtick pos=bottom,
            ytick pos=left,
            x tick label style={
                /pgf/number format/.cd,
                set thousands separator={\,}
            },
            xtick distance=500,
            enlarge x limits=0.03,
            restrict x to domain=0:2500,
            ymin=74.5,
            ymax=83,
        ]
            \addplot+[
                mark=*,
                color=plotcolour1,
                mark options={fill=plotcolour1, draw=plotcolour1!90!black, solid},
                thick,
            ] table [x=num_samples, y expr=\thisrow{mean_comet}*100, col sep=comma] {every100-qwen3.5-4b-no-thinking.csv};
            \addlegendentry{Qwen3.5 4B [I:\cmark]}
            
            \addplot+[
                mark=o,
                color=plotcolour1,
                mark options={fill=plotcolour1, draw=plotcolour1!90!black, solid},
                dashed,
                thick,
            ] table [x=num_samples, y expr=\thisrow{mean_comet}*100, col sep=comma] {every100-qwen3.5-4b-no-thinking-enable_thinking=False.csv};
            \addlegendentry{Qwen3.5 4B [I:\xmark]}
            
            \addplot+[
                mark=*,
                color=plotcolour2,
                mark options={fill=plotcolour2, draw=plotcolour2!90!black, solid},
                thick,
            ] table [x=num_samples, y expr=\thisrow{mean_comet}*100, col sep=comma] {every100-qwen3.5-9b-no-thinking.csv};
            \addlegendentry{Qwen3.5 9B [I:\cmark]}
            
            \addplot+[
                mark=o,
                color=plotcolour2,
                mark options={fill=plotcolour2, draw=plotcolour2!90!black, solid},
                dashed,
                thick,
            ] table [x=num_samples, y expr=\thisrow{mean_comet}*100, col sep=comma] {every100-qwen3.5-9b-no-thinking-enable_thinking=False.csv};
            \addlegendentry{Qwen3.5 9B [I:\xmark]}

        \end{axis}
        \end{tikzpicture}
    \caption{Translation quality across number of training samples.
        \textnormal{The COMET score for the models evaluated on a subset of 4\,000
        validation samples in intervals of 100 seen training samples. \textit{(left)}
        models trained with thinking, marked as [T:\cmark], and \textit{(right)}
        trained without thinking [T:\xmark]. Solid lines correspond to
        performing inference with thinking enabled [I:\cmark], whereas dashed
        lines indicate that thinking has been disable during the inference
        [I:\xmark].
        }
    }\label{fig:num-training-samples}
\end{figure*}
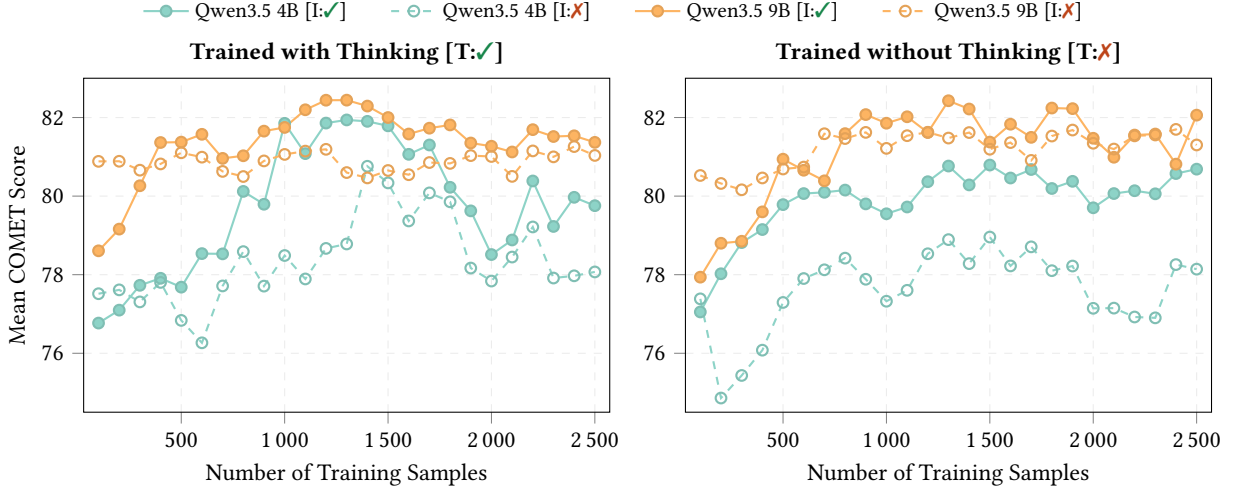

\section{Introduction}

Legal work runs on language, and increasingly that language is generated, analysed, and translated by machines. Large language models now assist across the legal workflow---from research and contract analysis to drafting, summarisation, and translation. Translation stands out among these applications: it is what makes legislation, judicial decisions, contracts, and regulatory texts accessible across language communities. The stakes are highest in multilingual jurisdictions such as Switzerland, where legal documents must be produced and interpreted with equal authority in German, French, and Italian.

Recent LLMs have markedly improved legal translation quality. On demanding benchmarks such as \verb|SwiLTra-Bench|~\cite{swiltra}, the Swiss Legal Translation Benchmark, reinforcement learning with verifiable rewards (RLVR) has been shown to push small open-weight models beyond what supervised fine-tuning (SFT) achieves~\cite{reasoning-before-translation}---though still short of the strongest frontier models.

These gains, however, surface a practical tension. The largest reasoning models tend to deliver the best translations, but they also demand far more computation, inference time, and money. For organisations translating thousands or millions of documents, chasing maximal accuracy may not be economically defensible. The question is no longer \emph{which model is most accurate}, but \emph{which model best balances legal reliability against inference cost}. 

This question of cost--quality tradeoffs is becoming increasingly central to the field. \citet{zhuang2025beyond} introduce a framework that provides a quantifiable decision-making tool for selecting the best AI technology within a limited budget, marking a key shift in the industry's focus from the pursuit of ever-larger models toward the efficiency of engineered, measurable application deployment. Reasoning exemplifies this tradeoff: while it has emerged as a promising approach for enhancing LLM capabilities across a wide range of tasks, generating explicit reasoning introduces additional output tokens, increasing inference cost and latency.

In this work, we study the cost-quality tradeoff for legal
machine translation, asking whether reasoning traces should be enabled during
\textit{training} and \textit{inference}. Our main findings are:

\begin{enumerate}
    \item Reasoning during both \textit{training} and \textit{inference}
    yields the best translation quality, at the cost of longer outputs
    and higher inference cost and latency.
    \item Mismatched configurations should be avoided: disabling
    reasoning at inference after reasoning-based training degrades
    quality below the reasoning-free baseline, while enabling it
    without reasoning-based training triples the inference cost.
\end{enumerate}

\section{Related Work}

\subsection{RLVR for Legal Translation}

Several recent studies have investigated the impact of RL on machine translation \citep{he2025r1,wang2026deep}. Recently, \cite{reasoning-before-translation} demonstrated that RLVR consistently outperforms SFT across different model sizes and language pairs for legal translation on the \verb|SwiLTra-Bench| benchmark.
They keep the rewards to a minimum, by using a structural reward to ensure that
the thinking and the translation are inside the appropriate XML tags to
facilitate their extraction from the response. The reward for the translation
quality uses chrF~\cite{chrf}, which can be computed efficiently and provides
continuous values in the range of $[0, 1]$ to have a nuanced distinction between
the sampled responses.

\subsection{The Economics of RLVR}

Given that RLVR improves translation quality, a natural question is how
to apply it most economically---in particular, whether explicit
reasoning is genuinely necessary. \citet{chen2026does} argue that RLVR
does not endow models with fundamentally new reasoning capabilities.
Instead, the observed gains arise from more effectively sampling
reasoning trajectories that are already present in the base model's
distribution: improvements in benchmark performance reflect better
exploitation of existing capabilities rather than the acquisition of
new ones.

If reasoning traces merely resurface what the base model already knows,
their cost deserves scrutiny. From an economic point of view, this
motivates two questions: 1)~is the reasoning step necessary during RL
training, given that generating responses with reasoning increases
training time and cost; and 2)~is reasoning necessary at inference
time, where the additional output tokens directly inflate latency and
cost?

\subsection{Energy and Inference Cost Estimation}

For frontier models accessed through commercial APIs, inference cost is
straightforward to compute: providers charge per token, with separate
rates for input and output tokens. 

For locally deployed models, however,
no such unified pricing exists, and the true cost depends on numerous
factors, including hardware, energy consumption, and infrastructure
overhead. \citet{zhuang2025beyond} propose a practical method for
translating task execution time into hardware cost. They first estimate
an hourly GPU cost as the sum of three components---depreciation, power
consumption, and maintenance---parametrised by the GPU purchase price,
depreciation period, utilisation rate, average power draw, data center
power usage effectiveness, electricity price, and maintenance rate.
According to \citet{zhuang2025beyond}, the total cost of an
inference workload is then obtained by multiplying this hourly rate by
the total execution time.

\begin{equation}
C_{\text{GPU,hour}} \approx C_{\text{depreciation}} + C_{\text{power}} + C_{\text{maintenance}}
\end{equation}

\section{Experiments}

To investigate the effectiveness of the reasoning in RLVR, we train models
with Group Relative Policy Optimisation (GRPO)~\cite{deepseek-math}, with and
without reasoning traces. This is achieved by providing a reward for the structure of the response where
the thinking process must be put inside the \texttt{<think></think>} tag while
the final translation is within the \texttt{<translation></translation>} tag,
such that it can be extracted easily. Disabling the reasoning is enforced by
requiring the thinking tag to be empty, and for efficiency reasons, the empty
tag is prefilled when sampling responses for the GRPO objective during training.
The reward for the translation quality uses the chrF score after
being extracted from the translation tag.

We fine-tune Qwen3.5~\cite{qwen3}, specifically the 4B and 9B parameter
variants, on the \verb|SwiLTra-Bench| dataset using QLoRA~\cite{qlora} with a
group size of $G = 4$, i.e.\ four different responses are sampled for each input
sentence.

As the cost effectiveness is one of our topics of interest, we save a
checkpoint of the model after every 100 training samples in order to assess
the impact of the amount of training data, and therefore training time, in
relation to the translation quality. In order to minimise training and inference
times we resort to optimised libraries, namely
\texttt{unsloth}\footnote{\url{https://github.com/unslothai/unsloth}} for
training and vLLM~\cite{vllm} for inference.\footnote{The code used for the experiments is available in the GitHub repository \href{https://github.com/aixiuxiuxiu/Legal-MT-SFT-RL}{https://github.com/aixiuxiuxiu/Legal-MT-SFT-RL}}

\begin{table*}[ht]
    \caption{Result for the best models on the SwiLTra-Bench test set. 
    \textnormal{Each model has been trained with (\cmark) and without (\xmark) thinking and subsequently evaluated by enabling or disabling thinking during inference. The estimated costs for the training and the inference of the complete test set are given alongside the number of generated tokens to make the overhead of the thinking quantifiable.
    }} \label{tab:results-large}
    \centering
    \begin{tabular}{l|cc|cccc|cc|r}
        \toprule
         & \multicolumn{2}{c|}{Thinking} & \multicolumn{4}{c|}{Metrics} & \multicolumn{2}{c|}{Cost} & Output \\
        \multicolumn{1}{c|}{Model} & Training & Inference & chrF $\uparrow$ & COMET $\uparrow$ & METEOR $\uparrow$ & MetricX $\downarrow$ & Training & Inference & Tokens \\
        \midrule
        \multirow{4}{*}{Qwen3.5 4B}
       & \xmark & \xmark & 53.79 \textcolor{faded}{$\pm$ 0.10} & 79.09 \textcolor{faded}{$\pm$ 0.07} & 50.11 \textcolor{faded}{$\pm$ 0.12} & 4.94 \textcolor{faded}{$\pm$ 0.02} & \multirow{2}{*}{\$5.46} & \$0.07 & 1.08M \\ 
       & \xmark & \cmark & 55.54 \textcolor{faded}{$\pm$ 0.10} & 81.05 \textcolor{faded}{$\pm$ 0.07} & 52.35 \textcolor{faded}{$\pm$ 0.12} & 4.18 \textcolor{faded}{$\pm$ 0.02} & & \$0.86 & 19.85M \\
       & \cmark & \xmark & 52.69 \textcolor{faded}{$\pm$ 0.11} & 78.61 \textcolor{faded}{$\pm$ 0.08} & 48.14 \textcolor{faded}{$\pm$ 0.13} & 5.01 \textcolor{faded}{$\pm$ 0.02} & \multirow{2}{*}{\$12.56} & \$0.10 & 1.90M \\ 
       & \cmark & \cmark & 56.33 \textcolor{faded}{$\pm$ 0.10} & 82.05 \textcolor{faded}{$\pm$ 0.06} & 52.13 \textcolor{faded}{$\pm$ 0.12} & 3.91 \textcolor{faded}{$\pm$ 0.02} & & \$0.24 & 8.08M \\
        \midrule
        \multirow{4}{*}{Qwen3.5 9B}
       & \xmark & \xmark & 55.81 \textcolor{faded}{$\pm$ 0.10} & 81.66 \textcolor{faded}{$\pm$ 0.06} & 53.42 \textcolor{faded}{$\pm$ 0.12} & 4.03 \textcolor{faded}{$\pm$ 0.02} & \multirow{2}{*}{\$4.54} & \$0.09 & 1.05M \\
        &  \xmark & \cmark & 56.90 \textcolor{faded}{$\pm$ 0.12} & 82.32 \textcolor{faded}{$\pm$ 0.08} & 54.48 \textcolor{faded}{$\pm$ 0.14} & \textbf{3.73} \textcolor{faded}{$\pm$ 0.02} & & \$1.03 & 20.67M \\
        & \cmark & \xmark & 54.60 \textcolor{faded}{$\pm$ 0.10} & 80.80 \textcolor{faded}{$\pm$ 0.07} & 50.96 \textcolor{faded}{$\pm$ 0.13} & 4.19 \textcolor{faded}{$\pm$ 0.02} & \multirow{2}{*}{\$11.61} & \$0.19 & 2.00M  \\
        & \cmark & \cmark & \textbf{57.51} \textcolor{faded}{$\pm$ 0.10} & \textbf{82.50} \textcolor{faded}{$\pm$ 0.06} & \textbf{54.61} \textcolor{faded}{$\pm$ 0.12} & 3.78 \textcolor{faded}{$\pm$ 0.02} & & \$0.28 & 6.27M \\
        \bottomrule
    \end{tabular}
\end{table*}

\section{Results}

The translation quality is evaluated with the following four metrics: chrF~\cite{chrf}, METEOR~\cite{meteor}, COMET~\cite{comet}, and
MetricX~\cite{metricx}. We put most weight behind the COMET score, as
\citet{reasoning-before-translation} have found the strongest correlation with
the judgement of human experts on the \verb|SwiLTra-Bench| dataset.

Since the thinking can be turned on and off during training or inference
independently of each other, we compare all possible permutations and indicate
within brackets whether thinking was used during training (T) and during
inference (I) with a check mark (\cmark) if it was enabled or a cross (\xmark) if
disabled. For example, \mbox{[T:\cmark, I:\xmark]} means, the model was trained with
thinking but the inference was performed without thinking.

\subsection{Number of Training Samples}

\autoref{fig:num-training-samples} depicts the COMET score on a subset of 4\,000
validation samples for the models trained in increments of 100 training samples.
The left plot contains the models that have been trained with thinking enabled,
but instead of evaluating them only in the same setting, the
inference has equally been run with the thinking turned off (indicated by the
dashed lines). Similarly, the right plot evaluates both scenarios but for the
models that have been trained without thinking.

It can be observed that enabling thinking during inference has a positive
effect on translation quality, regardless of whether the model has been
trained with thinking. The difference is much more noticeable for the smaller,
4B parameter model. Importantly, training without thinking, which enforces an empty \texttt{<think></think>}, does not destroy the thinking ability that the base model exhibits, which is consistent with the effects observed in SFT vs. RL comparisons~\cite{sft-vs-rl, rule-based-rl-document-classification}.

For this particular dataset, between 1\,000 to 1\,500 translation pairs are adequate to reach a plateau. While increasing the number of training samples beyond the depicted range could improve the overall results, it is unlikely that the improvements are disproportionally greater than the additional computational demands. The same can be said for the extra tokens that need to be generated for the reasoning during training, hence it is important to take the costs into account.

\subsection{Cost-Quality Tradeoff}

Following \citet{zhuang2025beyond}, we estimate the running cost of the models
based on the total execution time with a standardised cost of \$0.79/hour. While
the real cost depends on the specific hardware as well as the electricity cost,
it gives a reasonable approximation.

For each of the trained models, we select the best checkpoints and run the
inference on the \verb|SwiLTra-Bench| test set, consisting of roughly 18.1k
sentence pairs. The results are summarised in \autoref{tab:results-large} including
the estimated costs for the training and inference.

When it comes to cost, the inference of Qwen3.5 9B is more expensive, as
expected due to the higher parameter count, with differences ranging from \$0.02 up to
\$0.17. The gap widens as the number of output tokens increases, which reveals the primary inference bottleneck of being limited by the memory bandwidth in the autoregressive generation process.

A considerable difference can be observed in the generated output tokens. When
the model is trained with thinking, [T:\cmark], the reasoning makes the
responses $2-3 \times$ longer, although it also boosts the translation
quality, such as the COMET score by $2-4$ percentage points. The starkest
difference however, is during inference with thinking enabled. When the models
are trained without thinking, the output tokens shoot up from 8.08M to 19.85M
\textit{(+145\%)} for Qwen3.5 4B, and from 6.27M to 20.67M \textit{(+230\%)} for
Qwen3.5 9B, compared to the same inference if they were trained with thinking.
These additional reasoning tokens do not result in a better translation and are
in fact off by slightly over one percentage point in regards to the COMET score.
Therefore, the substantial increase in inference cost is not justified.

The best result is achieved by Qwen3.5 9B [T:\cmark, I:\cmark], which
incorporates thinking into both the training and inference, with a COMET score
of $82.50$. 
Enabling thinking during inference consistently improves the translation
quality, regardless of whether the model has been trained to produce thinking
traces. Training with thinking leads to marginal improvements at
increased training cost. However, the
aforementioned difference in output tokens tips the scale firmly in favour of
training with thinking, whose most important effect seems to be taming the
thinking process.

Training is often considered a one-time fixed cost, as the same model is
expected to be used for many more inference requests across a longer time span,
which surpasses the training cost fairly quickly, hence the inference cost
becomes the deciding factor. \autoref{fig:cost-plot} shows the COMET scores of
all models in relation to the inference cost, including a comparison with
commercial models with the pricing for their API requests on the same test set.
The models on the Pareto frontier represent the best tradeoff between inference
cost and translation quality.

\pgfdeclareplotmark{localmark}{
    \pgfpathcircle{\pgfpointorigin}{\pgfplotmarksize}
    \pgfsetfillcolor{plotcolour7}
    \pgfsetstrokecolor{plotcolour7!70!black}
    \pgfusepathqfillstroke
}

\pgfdeclareplotmark{commmark}{
    \pgfpathcircle{\pgfpointorigin}{\pgfplotmarksize}
    \pgfsetfillcolor{plotcolour4}
    \pgfsetstrokecolor{plotcolour4!70!black}
    \pgfusepathqfillstroke
}

\begin{figure}[h]
\begin{tikzpicture}
    \begin{axis}[
        xlabel={Inference Cost (US\$)},
        ylabel={Mean COMET Score},
        grid=major,
        enlargelimits=0.15,
        grid=major,
        grid style={gray!15, dashed},
        tick align=outside,
        xtick pos=bottom,
        ytick pos=left,
        x tick label style={
            /pgf/number format/.cd,
            set thousands separator={\,}
        },
        log ticks with fixed point,
        xtick={0.1, 1, 10, 100},
        xminorticks=false,
        xmode=log,
        legend style={
            at={(0.5, 1.0046)},
            anchor=south, 
            legend columns=2,
            legend cell align={left},
            font=\small,
            draw=none,
            column sep=0.1cm,
            /tikz/every even column/.append style={column sep=0.5cm}
        },
    ]
    
        \addplot [
            scatter,
            only marks,
            color=plotcolour7,
            fill=plotcolour7,
            mark=localmark,
            visualization depends on={value \thisrow{parameters} \as \myparams},
            scatter/@pre marker code/.append style={
                /tikz/mark size={1.2 * ln(\myparams)}
            }
        ] table [x=cost_inference, y=comet, col sep=semicolon] {cost_qwen.csv};
        \addlegendentry{Local Models}
        
        \addplot [
            mark=none, 
            draw=none,
            forget plot, 
            point meta=explicit symbolic,
            nodes near coords,
            visualization depends on={value \thisrow{anchor} \as \myanchor},
            visualization depends on={value \thisrow{xshift} \as \myxshift},
            visualization depends on={value \thisrow{yshift} \as \myyshift},
            every node near coord/.style={
                font=\tiny,
                text=black,
                anchor=\myanchor, 
                xshift=\myxshift,
                yshift=\myyshift,
                inner sep=4pt
            }
        ] table [x=cost_inference, y=comet, meta=label, col sep=semicolon] {cost_qwen.csv};

        \addplot [
            color=black!70,
            densely dashed,
            mark=none
        ] table [x=cost_inference, y=comet, meta=label, col sep=semicolon] {cost_pareto.csv};
        \addlegendentry{Pareto Frontier}

        \addplot [
            scatter,
            only marks,
            color=plotcolour4,
            fill=plotcolour4,
            mark=commmark,
            mapped color/.style={draw=blue!70!black, fill=blue!50},
            visualization depends on={value \thisrow{params} \as \myparams},
            scatter/@pre marker code/.append style={
                /tikz/mark size={1.2 * ln(\myparams)}
            }
        ] table [x=cost, y=comet, col sep=comma] {cost_commercial.csv};
        \addlegendentry{Commercial Models}
        
        \addplot [
            mark=none, 
            draw=none,
            forget plot,
            point meta=explicit symbolic,
            nodes near coords,
            visualization depends on={value \thisrow{anchor} \as \myanchor},
            visualization depends on={value \thisrow{xshift} \as \myxshift},
            visualization depends on={value \thisrow{yshift} \as \myyshift},
            every node near coord/.style={
                font=\tiny,
                text=black,
                anchor=\myanchor, 
                xshift=\myxshift,
                yshift=\myyshift,
                inner sep=4pt
            }
        ] table [x=cost, y=comet, meta=model, col sep=comma] {cost_commercial.csv};
    \end{axis}
\end{tikzpicture}
\caption{Reasoning Pareto frontier. \textnormal{
Each point depicts the model's translation quality in relation to its inference
cost. The reasoning configuration is specified as
\mbox{[T:\cmark, I:\cmark]}, for whether thinking was employed during 
training and inference, respectively. The marker size is proportional to the
parameter count of the model.
}}\label{fig:cost-plot}
\end{figure}
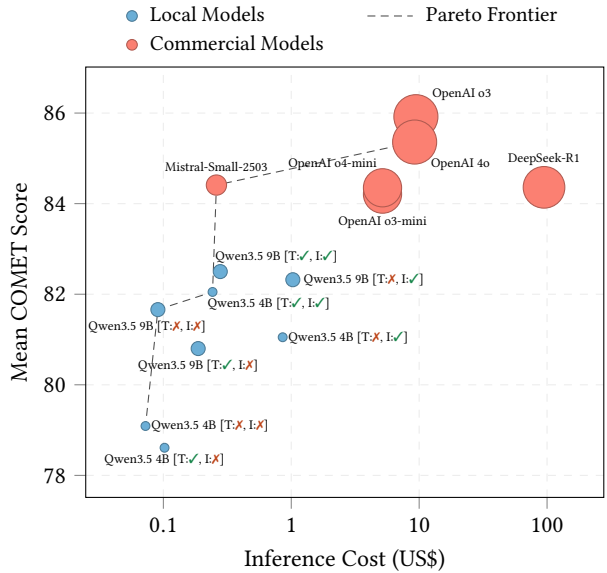

\section{Conclusion}

The experiments conducted in this paper showed that including the model's
reasoning in the output has a positive effect on the overall translation
quality. By adding the reasoning to the output, the model produces significantly
more tokens, which result in longer generation times and therefore increased
operating costs. To minimise the excess of tokens, the thinking needs to be
integrated into the training. Even though we did not impose any restrictions on
the reasoning traces, letting the model generate responses with reasoning for
the GRPO objective keeps them more compact, with a reduction in tokens of up to
70\% for Qwen3.5. The cost-quality tradeoff favours the models that
preserve the thinking mode between training and inference, i.e.\ either no
thinking at all or always thinking. 

While the training costs can be reduced by disabling the thinking mode
during training, the inference costs quickly surpasses the total costs if the thinking
is going to be enabled after the fact. A much more impactful lever is the number
of training samples used during the training. Within the context of machine
translation in the legal domain, or more specifically for the
\verb|SwiLTra-Bench| dataset, limiting the training to include around 1\,500 training
samples is sufficient to reach the plateau after which every small improvement
incurs a disproportionally large cost. Therefore it would likely be more
beneficial to curate a set of high quality training samples.

\section*{Acknowledgements}

The authors would like to thank the Hasler Foundation and the Mercator Foundation Switzerland for their financial support.

\bibliographystyle{ACM-Reference-Format}
\bibliography{references}

\end{document}